\documentclass[sigconf,nonacm]{acmart}
\usepackage{booktabs}
\usepackage{enumitem}

\AtBeginDocument{%
  }

\begin{document}

\title{Expert Evaluation of LLM's Open-Ended Legal Reasoning\\on the Japanese Bar Exam Writing Task}

\author{Jungmin Choi}
\affiliation{%
  \institution{RIKEN}\state{Tokyo}\country{Japan}}
\email{jungmin.choi@riken.jp}

\author{Keisuke Sakaguchi}
\affiliation{%
  \institution{Tohoku University}\state{Miyagi}\country{Japan} }
\email{keisuke.sakaguchi@tohoku.ac.jp}

\author{Hiroaki Yamada}
\affiliation{%
  \institution{Tokyo Metropolitan University}\state{Tokyo}\country{Japan}}
\email{ymd@tmu.ac.jp}

\renewcommand{\shortauthors}{Choi et al.}

\begin{abstract}
Large language models (LLMs) have shown strong performance on legal benchmarks, including multiple-choice components of bar exams. However, their capacity for generating open-ended legal reasoning in realistic scenarios remains insufficiently explored.
Notably, to our best knowledge, there are no prior studies or datasets addressing this issue in the Japanese context.

This study presents the first dataset designed to evaluate the open-ended legal reasoning performance of LLMs within the Japanese jurisdiction. The dataset is based on the writing component of the Japanese bar examination, which requires examinees to identify multiple legal issues from long narratives and to construct structured legal arguments in free text format.
Our key contribution is the manual evaluation of LLMs' generated responses by legal experts, which reveals limitations and challenges in legal reasoning. Moreover, we conducted a manual analysis of hallucinations to characterize when and how the models introduce content not supported by precedent or law.

Our real exam questions, model-generated responses, and expert evaluations reveal the milestones of current LLMs in the Japanese legal domain. Our dataset and relevant resources will be available online.
\end{abstract}




\maketitle

\section{Introduction}

Large language models (LLMs) have recently achieved strong results on a variety of LegalNLP benchmarks, many of which take the form of classification or short-form question answering (e.g., entailment-style tasks, label prediction, and short retrieval-based QA)~\citep{chalkidis-etal-2022-lexglue,guha2023legalbench}.
Such gains have fueled interest in deploying LLMs in high-stakes legal workflows.

However, practical legal reasoning rarely consists of selecting a predefined option.
This process involves identifying issues, articulating applicable laws or norms, applying statutes and precedents to facts, and justifying conclusions in a structured written argumentation.
Furthermore, LLMs may generate legally or factually inaccurate or unsupported claims.
Taken together, these considerations motivate the development of benchmarks that (i) require free-form legal argumentation over complex fact patterns and (ii) provide expert-grounded evaluation beyond a single overall score.

Professional qualification examinations provide a natural source of evaluation tasks because they are explicitly designed to assess application of laws under realistic scenarios.
Recent work has used bar exams to assess frontier LLMs in the U.S.\ context, reporting strong performance on the Uniform Bar Exam, including its multiple-choice and essay writing components~\citep{katz2023gpt4bar,openai2023gpt4}.
However, they do not address open-ended written reasoning under controlled prompting interventions or in the Japanese jurisdiction.
Figure~\ref{fig:bar-exam-example} shows a translated excerpt of a Japanese bar exam writing-test question, illustrating the long-form fact narrative and open-ended prompt that require structured legal analysis rather than option selection.

In Japan, entry into legal professions, such as judges, public prosecutors, and attorneys, is nationally regulated through this examination.
Passing the examination is mandatory for admission to the unified post-exam judicial training program administered by the Legal Training and Research Institute of the Supreme Court.
The examination consists of a multiple-choice test component and a writing test component covering core legal areas such as Civil Code, Penal Code, their corresponding procedural codes, the Constitution of Japan, Commercial Code, and administrative laws.
The writing component requires examinees to analyze factually complex prompts by identifying relevant legal issues, interpreting and applying statutory provisions and precedents, and producing coherent, structured arguments.
Thus, this examination provides a suitable foundation for evaluating open-ended legal reasoning that current benchmarks often fail to capture.

We introduce a new dataset derived from the writing test component of the Japanese bar exam.
Each instance in the dataset consists of an exam question, a response generated by an LLM, and detailed manual evaluations by legal experts.
The expert evaluation assesses model outputs across several dimensions, including legal correctness, identification of issues, and argumentative structure.
Additionally, we conduct a post-hoc analysis of hallucinations, systematically identifying instances where models introduce content that is not supported by applicable laws or precedents.
Overall, by combining real exam questions, model-generated answers, and expert assessments, our dataset enables a detailed and realistic evaluation of open-ended legal reasoning.

Our contributions are threefold.
1) We constructed a dataset comprising the Japanese bar exam writing tests and their answers generated by LLMs. 
2) We perform manual evaluation of model outputs with expert legal annotators, providing fine-grained assessments of legal correctness, reasoning quality, and argumentative coherence.
3) We conduct a hallucination analysis on the outputs from models, offering insight into the types of failure modes that present in open-ended legal generation tasks.
\begin{figure}[t]
\centering
\fbox{%
\begin{minipage}{0.95\linewidth}
\small
\textbf{Facts}\\
1.~A operated a sole proprietorship using a single truck (hereinafter, the ``Truck'') to purchase fresh produce from growers and sell it to retailers and restaurants.\\
2.~On September 10, 2017, A entered into a contract with B to purchase 5 kilograms of matsutake mushrooms for a price of 500{,}000 yen (hereinafter, the ``Sales Contract''). Under the Sales Contract, delivery of the matsutake was to be made on the night of September 21, 2017 at B's warehouse (the `` Warehouse'') located near B's apple orchard, \emph{concurrently with payment} (i.e., delivery against payment).\\
\vdots

\textbf{Question 1 (prompt)}\\
Assuming Facts 1 through 9 above, is B's claim against A for payment of the purchase price under the Sales Contract (as stated in Fact 9) legally justified? Answer with reasons.
\end{minipage}%
}
\caption{Excerpt from a writing-test question, machine-translated and manually corrected. This subquestion is about 700–800 words in English. }
\label{fig:bar-exam-example}
\end{figure}
\section{Related Work}
\label{sec:related}

Various datasets and benchmarks have been introduced in the domain of legal NLP, including information retrieval, legal textual entailment, sentence classification, and question answering over statutes and case law. With the rise of LLMs, the trend has shifted from classification-style evaluation (e.g., selecting a label) toward assessing generation-style legal reasoning (e.g., producing structured arguments). 
Our dataset lies at the intersection of them, targeting realistic, open-ended legal analysis, and supports expert-grounded evaluation of model outputs.

\subsection{Legal Benchmark Datasets}
There are major benchmarks employing various tasks in legal language understanding and reasoning. LexGLUE~\citep{chalkidis-etal-2022-lexglue} is a benchmark suite comprising multiple English legal datasets and tasks, enabling systematic comparisons across domains such as case law, legislation, and contracts. 
More recently, LegalBench~\citep{guha2023legalbench} extends the scope of evaluation toward legal reasoning, reflecting the emergence of LLMs by compiling a large set of tasks constructed with legal expertise, with the explicit goal of measuring legally meaningful reasoning skills rather than narrow text classification accuracy.

Another line of work introduces tasks from legal education and professional qualification exams. Such sources are attractive because they are intentionally designed to assess whether examinees can arrange facts and apply laws to them. For example, JEC-QA~\citep{zhong2019jecqa} is constructed from China's national judicial examination, and \citet{choi2023gpt_japanese_bar_jp} similarly evaluates GPT-family models on the multiple-choice component of the Japanese bar examination.
In the common-law context, CaseHOLD~\citep{zheng2021doespretraininghelpassessing} provides multiple-choice questions derived from judicial opinions, focusing on identifying the correct holding given a scenario. 
These datasets are useful for benchmarking legal knowledge and the retrieval ability of precedents and statutes. However, their answer formats often constrain evaluation to selection among predefined options, making it difficult to assess whether a model can produce coherent legal arguments.

Shared tasks also play an important role in the legal domain. The COLIEE competition includes tasks such as case law retrieval and statute-based entailment~\citep{rabelo2020coliee}. COLIEE demonstrated how real exams can be utilized in standardized evaluation tasks. However, their primary evaluation settings remain classification-based evaluations for retrieval and entailment performance, which do not require free-form legal argumentation.

Given the limitations of multiple-choice and short-answer formats, there are datasets designed to assess the long-form answers and reasoning. LLeQA~\citep{10.1609/aaai.v38i20.30232} provides a long-form legal question answering task with sample answers written by experts. LEXam~\citep{fan2025lexambenchmarkinglegalreasoning} utilizes exam questions in the law domain. They include open-ended questions that require structured argumentation based on multi-step legal reasoning. GreekBarBench~\citep{chlapanis-etal-2025-greekbarbench} aligns closely with our work on motivation. They utilize bar exam style questions and provide scores of LLM-generated answers by human experts, focusing on legal analysis beyond surface correctness.

In spite of these advances, manual evaluations by experts for model-generated answers remain limited. Many datasets provide (i) questions along with reference answers, or (ii) scores on final outcomes for automatic scoring. However, they do not provide an expert assessment. This gap prevents us from analyzing where models fail in legal reasoning and intermediate steps.

\subsection{Hallucinations}
Reliability is the major concern in applying LLMs in the legal domain.
Legal problems are highly sensitive to factual details and legal interpretations. LLMs are known to produce fluent but unsupported arguments. Recent work reports the prevalence of legal hallucinations and their risks, including false citations and confidently stated but incorrect legal claims~\citep{Dahl_2024}.
The hallucination problems are reported even in legal domain-specific research tools.
\citet{magesh2024hallucinationfree} report that even popular RAG-based systems suffer from hallucination at considerable rates (17-33\%).
These issues motivate evaluation frameworks that explicitly separate reasoning quality and legal soundness from hallucination, rather than conflating these aspects into a single overall score.

\subsection{Our Dataset}
Our dataset has advantages beyond the previous datasets in three ways.
First, our dataset provides a task, targeting open-ended, exam-style legal reasoning in a setting that naturally elicits structured argumentation rather than simple classification tasks.
Second, our dataset provides pairs of questions and their LLM-generated answers, enabling analysis of model behavior in legal reasoning.
Third, our dataset additionally provides expert evaluation results. They include not only overall scoring about legal reasoning, but also hallucinations.
Our dataset enables a fine-grained empirical study of LLMs' legal reasoning capabilities and failures in a high-stakes, professionally aligned evaluation setting.

\section{Dataset Construction}
\subsection{Source}
Our dataset is constructed from official questions of the writing test of the Japanese National Bar Examination for the years 2017--2023. For each year, we include all subjects published in the exam, yielding 161 questions in total (23 per year).

In the writing test, each subject typically contains two to three independent questions.
Each question presents a detailed factual scenario as a prompt, and poses one or more subquestions that ask the examinee to identify and analyze the legally relevant issues and articulate reasoned views on the applicable law.
These questions are designed to elicit practical legal reasoning: examinees must select and interpret statutory provisions and precedents, distinguish among competing interpretations, and organize their analysis into coherent, well-structured arguments.

\subsection{Data Scheme}
We provide a structured representation of the writing test questions for reproducible evaluation and controlled prompting.
Each instance corresponds to a single question and contains question data and statutory references.

Question data comprises (i) a fact-rich narrative prompt, (ii) one or more subquestions, and (iii) any accompanying materials referenced in the prompt (e.g., statutory excerpts, tables/figures).

Statutory references are a collection of relevant statutes not explicitly cited in the questions.
One legally trained author manually compiled them with reference to third-party answers.
Each entry contains the law identifier, law name, and the full text of the referenced article in a lawtext-like\footnote{\url{https://github.com/yamachig/lawtext}} format.

\section{Experimental Settings}
\subsection{Models and Parameters}
We evaluate several state-of-the-art large language models representing different model families
and training paradigms.
Specifically, our experiments include the following models: \textbf{GPT-4o} (\texttt{gpt-4o-2024-11-20})~\citep{openai_gpt4o_syscard,openai_api_gpt4o_snapshot}, \textbf{o3} (\texttt{o3-2025-04-16})~\citep{openai_api_o3}, \textbf{Claude 3 Opus} (\texttt{claude-3-opus-20240229})~\citep{anthropic_claude3_modelcard}.

Unless otherwise specified, all models were evaluated using their default inference and decoding
configurations.
GPT-4o was run with standard sampling (temperature = 1.0, top-$p$ = 1.0).
The o3 model does not provide explicit sampling parameters for users.
For the reasoning effort, we used a default of medium.
Unless otherwise stated, no explicit maximum output token limit was imposed.

For all models, system prompts assign the role of a Japanese legal expert and require that outputs be based on Japanese law and written in a casual form (non-polite form)

Each question instance is input as a part of a user prompt.
The question statement is presented without modification, following an instruction specifying the temporal scope of applicable law (e.g., ``Answer based on laws and regulations in force as of [the date of the examination]'').
This instruction ensures that models reason within a consistent legal temporal constraint.

\subsection{Approaches}
In the experiments, we prepare three different settings for each LLM.
\textbf{ZS}: The zero-shot (ZS) setting is the most straightforward approach, involving no specialized prompts or in-context learning configurations. Models receive only the target questions as input.
\textbf{FS}: The few-shot (FS) employs in-context learning. The prompt includes example bar exam problems and corresponding sample answers.
These examples were drawn from the 2018 and 2019 exams, which do not overlap with our evaluation set used in the experiment.
\textbf{FS+Law}: Building on FS, this setting supplements the prompt with relevant legal articles, sourced from statutory references from our dataset.

\subsection{Data}
All experiments were conducted using bar exam questions from 2021, 2022, and 2023. From the full dataset described in Section~3, only questions from the seven compulsory subjects were utilized: The Constitution of Japan, Civil Code, Penal Code, administrative laws, Code of Civil Procedure, Code of Criminal Procedure, and commercial laws.
In total, we evaluate $21$ questions (7 per year across 2021--2023), producing $189$ model-generated answers under three prompting settings.

\subsection{Evaluation Metrics}
Our experiments relied on expert manual evaluation to assess outputs involving complex, extended legal reasoning.
The evaluation was conducted by legal experts from the faculty of law at a Japanese university. Each expert was assigned to a specific subject, for a total of 7 evaluators.
Experts assessed answers holistically, considering issue identification, legal analysis, application of law to facts, logical structure, and clarity of writing, in line with the written exam scoring philosophy. They could also optionally leave free-form comments.

The outputs were assessed using both categorical gradings and numerical scores.
The categorical gradings consist of four categories that generally correspond to the predefined score ranges in the bar exam grading policy: \textit{Excellent} (100--75), \textit{Good} (74--58), \textit{Adequate} (57--42), and \textit{Poor} (41--0).
We treat 25 as an important threshold: in the real exam, a score below 25\% is an instant failure, so an answer scoring at least 25 meets the minimum requirement.

\section{Results and Discussion}
\subsection{Overall Performance}
\begin{table*}[t!]
\small
\centering
\caption{Expert evaluation results}
\label{tab:results-main}
\begin{tabular}{@{}llrrrrrrrrrrrr@{}}
\toprule
\multicolumn{2}{c}{Approaches} & \multicolumn{8}{c}{Scores by subjects, averaged} & \multicolumn{4}{c}{Categorical grades for each setting}\\
\cmidrule(rl){1-2} \cmidrule(rl){3-10} \cmidrule(rl){11-14}
\multicolumn{1}{c}{Model} &
\multicolumn{1}{c}{Prompting} &
\multicolumn{1}{c}{Penal} &
\multicolumn{1}{c}{Criminal Proc.} &
\multicolumn{1}{c}{Const.} &
\multicolumn{1}{c}{Admin.} &
\multicolumn{1}{c}{Civil} &
\multicolumn{1}{c}{Comm.} &
\multicolumn{1}{c}{Civil Proc.} &
\multicolumn{1}{c}{Avg.} &
\multicolumn{1}{c}{Exc.} &
\multicolumn{1}{c}{Good} &
\multicolumn{1}{c}{Adeq.} &
\multicolumn{1}{c}{Poor} \\
\midrule
o3     & ZS     & 18.7 & 31.3 & 51.3 & 34.7 & 10.7 & 40.7 & 21.3 & 29.8 & 0 & 3 & 0 & 18 \\
o3     & FS     & 20.3 & 26.3 & 44.0 & 35.3 & 14.0 & 45.0 & 21.0 & 29.4 & 0 & 2 & 1 & 18 \\
o3     & FS+Law & 20.3 & 28.7 & 50.7 & 33.7 & 18.3 & 39.0 & 22.7 & 30.5 & 0 & 1 & 3 & 17 \\
GPT-4o & ZS     &  7.0 & 24.0 & 17.0 & 23.3 &  0.7 & 26.3 & 15.0 & 16.2 & 0 & 0 & 0 & 21 \\
GPT-4o & FS     &  9.7 & 16.7 & 23.3 & 18.3 &  2.7 & 28.3 & 18.3 & 16.8 & 0 & 0 & 0 & 21 \\
GPT-4o & FS+Law &  9.0 & 17.3 & 25.0 & 22.3 &  5.3 & 26.7 & 17.7 & 17.6 & 0 & 0 & 0 & 21 \\
Claude 3 Opus & ZS     & 18.3 & 31.3 & 20.7 & 25.7 &  7.3 & 33.7 & 19.0 & 22.3 & 0 & 0 & 0 & 21 \\
Claude 3 Opus & FS     & 26.3 & 30.7 & 50.7 & 39.0 &  9.3 & 32.0 & 24.7 & 30.4 & 0 & 1 & 4 & 16 \\
Claude 3 Opus & FS+Law & 31.3 & 38.7 & 49.7 & 51.0 & 14.7 & 47.0 & 27.3 & 37.1 & 0 & 3 & 9 &  9 \\
\bottomrule
\end{tabular}
\end{table*}

The left side of Table~\ref{tab:results-main} shows scores from each approach.
The scores are averages of scores over the three years.
The right side of Table~\ref{tab:results-main} shows the number of answers that received the corresponding categorical grade for each approach.
Across all answers, the average score is 25.6 out of 100, and only $9.0\%$ of answers are \emph{Adequate}; $5.3\%$ reach \emph{Good}. No answer is labeled as \emph{Excellent}.

When looking at the overall average scores, providing expert-curated statutes (FS+Law) improves outcomes compared to zero-shot (ZS) and few-shot without statutes (FS). However, the magnitude of the effect differs across models.
Claude 3 Opus benefits substantially from FS+Law, increasing its score from $30.4\%$ in FS to $37.1\%$, whereas GPT-4o consistently performs \emph{Poor}ly across all settings.
Scores from o3 are relatively stable across prompting settings, though its FS+Law still shows a better score.

\subsection{Expert Comments Analysis}
We analyze the expert comments available for $104/189$ answers.
Among the 104 answers, 36, 32, and 36 answers are from ZS, FS, and FS+Law promptings, respectively.
We summarize and report the remarkable comments, focusing on recurring keywords and phrases related to errors.

There are comments about the inappropriate format of answers, such as bullet fragments. These types of comments are more common in ZS (found in 10 out of 36 answers, $27.8\%$) than in FS ($2/32$, $6.3\%$) or FS+Law ($2/36$, $5.6\%$).
In contrast, comments indicating weak fact-sensitive application remain common across settings: ZS ($13/36$, $36.1\%$), FS ($10/32$, $31.3\%$), FS+Law ($10/36$, $27.8\%$).
These observations suggest that in-context learning enhances consistency in answer style compliance; however, it does not necessarily improve the mapping of complex facts to legal requirements during practical reasoning.

Expert comments noted missing or incorrect mentions of statutes. This type of comment decreased in FS+Law ($6/36$, $16.7\%$) from FS ($11/32$, $34.3\%$).
Providing gold statutes reduces statute-selection or citation errors, as we intended.
Moreover, there are 7 comments on answers from o3 models regarding hallucinations of legal resources, such as laws and precedents.

\subsection{Hallucination Check}

\begin{table}[t]
\small
\centering
\caption{Hallucinations by Subjects}
\label{tab:hallcinantion-subjects}
\begin{tabular}{@{}lrrr@{}}
\toprule
\multicolumn{1}{c}{Subject} & \multicolumn{1}{c}{Hallucination} & \multicolumn{1}{c}{Not hallucination} & \multicolumn{1}{c}{Sum} \\ \midrule
Civil    & 133 (24.7\%) & 406 (75.3\%)   & 539 (100\%)   \\
Commercial    & 117 (24.8\%) & 354 (75.2\%)   & 471 (100\%)   \\
Criminal Proc. & 115 (41.4\%) & 163 (58.6\%)   & 278 (100\%)   \\
Civil Proc. & 88 (21.2\%)  & 328 (78.8\%)   & 416 (100\%)   \\
Constitution    & 84 (20.6\%) & 323 (79.4\%)   & 407 (100\%)   \\
Penal    & 72 (16.0\%)  & 379 (84.0\%)   & 451 (100\%)   \\
Administrative   & 67 (14.8\%)  & 387 (85.2\%)   & 454 (100\%)  \\ \bottomrule
\end{tabular}
\end{table}

\begin{table}[t]
\small
\centering
\caption{Hallucinations by Approaches}
\label{tab:hallcinantion-models}
\begin{tabular}{@{}llrr@{}}
\toprule
\multicolumn{1}{c}{Model} & \multicolumn{1}{c}{Prompting} & \multicolumn{1}{c}{Hallucination} & \multicolumn{1}{c}{Not hallucination} \\ \midrule
o3            & ZS     & 236 (34.7\%) & 445 (65.3\%) \\
o3            & FS     & 177 (30.8\%) & 397 (69.2\%) \\
o3            & FS+Law & 126 (22.8\%) & 426 (77.2\%) \\
GPT-4o        & ZS     & 12 (11.7\%)  & 91 (88.3\%)  \\
GPT-4o        & FS     & 16 (11.5\%)  & 123 (88.5\%) \\
GPT-4o        & FS+Law & 17 (10.2\%)  & 149 (89.8\%) \\
Claude 3 Opus & ZS     & 21 (14.6\%)  & 123 (85.4\%) \\
Claude 3 Opus & FS     & 53 (17.6\%)  & 248 (82.4\%) \\
Claude 3 Opus & FS+Law & 18 \,\,(5.1\%)   & 338 (94.9\%) \\
\bottomrule
\end{tabular}
\end{table}

To systematically assess the extent to which the LLMs generate hallucinated legal content, we performed a post-hoc analysis on the hallucinated content. 
The analysis focuses on hallucinations involving laws and precedents, as they are principal components of legal reasoning and can be clearly distinguished.

We employed a law school graduate of a Japanese university to annotate the hallucinated, false laws and precedents. The annotator was instructed to first identify every mention of laws or precedents, and classify each as either a hallucination or not.
If a mention consisted solely of a name or a number identifier of a law or precedent, and if the mention is inconsistent with actual laws or precedents, the annotator labeled the mention as a hallucination. 
If a mention included a quotation of a part of or a whole law article or a precedent, and if its quoted content contradicted its original source, the annotator flagged the mention as hallucination.

We used all of 189 answers from the LLMs for the analysis. As a result, there are 2,575 mentions of laws and 441 mentions of precedents in total. Out of them, 381 mentions of laws and 295 mentions of precedents are hallucinations.
The hallucination rate for precedents (66.9\%) is substantially higher than that for laws (14.8\%), indicating that the model's knowledge is less reliable for precedents than for law articles.

Table~\ref{tab:hallcinantion-subjects} presents the distribution of hallucinations by subject.
The Civil Code showed the highest number of hallucinated mentions (133), while Administrative laws had the lowest (67). 
The total number of mentions to laws and precedents for the Code of Criminal Procedure was the lowest among all subjects (278), and its hallucination rate was the highest (41.4\%).
The LLMs are less likely to reference legal resources in the subject and more likely to make incorrect citations.

Comparing the results across models (Table~\ref{tab:hallcinantion-models}), approaches with o3 resulted in more frequent hallucinations than approaches with the other models. The differences in prompting also impacted the hallucinations. FS+Law consistently showed lower hallucination rates than those with FS or ZS across all the models.

\section{Conclusion}
We introduced a dataset for open-ended legal reasoning based on the writing test of the Japanese bar exam.
Using a subset, we manually evaluated 189 LLM-generated answers.
Performance remained limited.
The average score stayed at 25.6/100.
These results reveal a substantial gap between prior LegalNLP benchmark performance and the ability to produce legally coherent, well-structured arguments.
Furthermore, our analysis of hallucinated mentions of legal resources highlights substantial room for improvement in Japanese legal reasoning.

\section{Ethical Considerations and Limitations}
We summarize key ethical considerations and practical limitations of our dataset construction and evaluation protocol.
\begin{itemize}[leftmargin=*,noitemsep]
\item Expert grading and hallucination annotation were conducted by human evaluators who have expert knowledge in the Japanese legal domain. Annotators were compensated for their time.
\item There are limitations of reliability in expert scoring. Each subject was evaluated by a single expert. This design matches practical constraints (domain expertise, financial and time cost); however, it limits the extent to which we can quantify inter-rater variability. Also, we cannot perfectly replicate the true scoring protocol in the real exam. Thus, one should be careful when comparing our results with actual human examinees' scores and statistics.
\item We evaluated answers generated by specific API snapshots of GPT-4o, o3, and Claude 3 Opus. Because proprietary models evolve over time, results may not transfer directly to future versions, open-source models, or systems with different tool access.
We mitigate this by reporting the exact model identifiers and settings, but readers should interpret the findings as a snapshot of model behavior under the specified experimental conditions.

\end{itemize}

\section*{Acknowledgments}
This work was supported by JST PRESTO Grant Number JPMJPR236B.

\bibliographystyle{ACM-Reference-Format}
\bibliography{custom}
\appendix
\end{document}